\documentclass[11pt,letterpaper]{article}
\usepackage[margin=1in]{geometry}
\usepackage[T1]{fontenc}
\usepackage[utf8]{inputenc}
\usepackage{lmodern,microtype}
\ifdefined\pdfgentounicode
\input{glyphtounicode}
\fi
\usepackage{amsmath,amssymb,graphicx,booktabs,tabularx,longtable,array}
\usepackage{caption,placeins,xcolor}
\usepackage[round,authoryear]{natbib}
\usepackage[hyphens]{url}
\usepackage[colorlinks=true,linkcolor=blue!45!black,citecolor=blue!45!black,urlcolor=blue!45!black]{hyperref}
\title{Diagnosing Model--Item Interactions in Language Model Benchmarks with LSIRM}
\author{Jinwen Luo$^{1}$, Han Jiang$^{2}$, Ziang Xiao$^{2}$,\\
Minjeong Jeon$^{1}$, and Susu Zhang$^{3}$\\[4pt]
\small $^{1}$University of California, Los Angeles\\
\small $^{2}$Johns Hopkins University\\
\small $^{3}$University of Illinois Urbana-Champaign}
\date{}
\begin{document}
\maketitle
\begin{abstract}
Language models with similar benchmark accuracy can succeed on different items. Overall scores do not show where their performance differs or which item properties might explain those differences. We demonstrate how the latent space item response model (LSIRM) can help evaluators investigate these patterns and their consequences for model comparison. We analyze 1,348 language models on 5,761 BBH items and compare ways of presenting and scoring answers across BBH, GPQA, MMLU-Pro, and MuSR. We examine LSIRM maps, which summarize differences in performance across models and items, alongside item properties and records of correct and incorrect responses. In BBH, nearby items on these maps often share the same correct answer, such as ``Yes'' or ``No.'' Across the four benchmarks, this pattern is stronger when answers are scored using option labels, such as A, B, C, and D, than when using the text of the options. We then estimate each model's tendency to perform better or worse on items with particular correct answers and quantify how this tendency contributes to its score. This separates an estimated contribution associated with correct answers from the remaining differences in item performance. Two models' total numbers of correct answers differed by only 14 out of 5,761 items. Excluding the estimated answer-category contribution from their scores moved their ranks hundreds of places in opposite directions, while keeping all other estimates unchanged. The approach helps evaluators identify differences hidden by overall accuracy and understand how the choice of what to include in a score affects model rankings.
\end{abstract}

\section{Introduction}

\begin{figure}[t]
    \centering
\includegraphics[width=\linewidth]{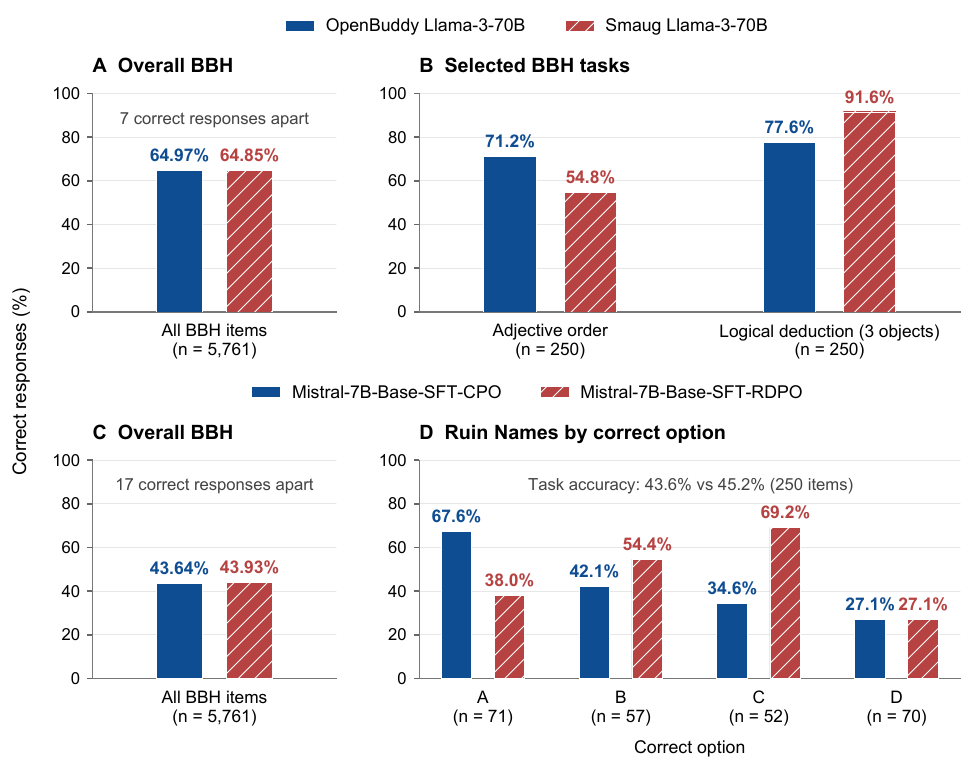}
\caption{ \textbf{Similar overall accuracy conceals differences across tasks and correct-answer groups.} (A--B) OpenBuddy and Smaug, two 70B models, differ by seven correct responses across 5,761 BBH items, yet their ordering reverses between adjective order (Hyperbaton) and logical deduction with three objects (250 items per task). (C--D) The Mistral 7B Base-SFT-CPO and Base-SFT-RDPO models differ by 17 correct responses overall and four on Ruin Names (109 versus 113 out of 250), yet their ordering reverses between items whose correct answers are A and C. Bars show observed accuracy; $n$ denotes the number of items. In D, A--D group items by the correct option in the answer key. The panels present selected descriptive examples from the same BBH response matrix; full model identifiers appear in Table~\ref{tab:s-example-identities}. }
    \label{fig:introduction-comparisons}
\end{figure}

In the Hugging Face Open LLM Leaderboard response data analyzed here \citep{openllmv2}, OpenBuddy Llama-3-70B and Smaug Llama-3-70B achieve accuracies of 64.97\% and 64.85\% on BIG-Bench Hard (BBH) \citep{bbh2022}. Their totals differ by seven correct answers out of 5,761 items. Yet OpenBuddy performs better on questions about adjective order, whereas Smaug performs better on logical deduction involving three objects (Figure~\ref{fig:introduction-comparisons}A--B). Their similar overall accuracy therefore conceals differences in where they succeed and fail. These differences may matter when choosing a model for particular tasks.

Differences also appear within the same task. Two Mistral 7B variants, Base-SFT-CPO and Base-SFT-RDPO, differ by 17 correct answers across BBH and by four on the 250 Ruin Names items, which ask models to identify humorous edits to names. CPO performs better on items whose correct answer is A, while RDPO performs better when C is correct, with gaps of approximately 30 and 35 percentage points, respectively (Figure~\ref{fig:introduction-comparisons}C--D). Even a task-level average can conceal different patterns of success across its items.

We use the latent space item response model (LSIRM) \citep{jeon2021} as a diagnostic tool to examine these model--item differences across a BBH cohort of 1,348 language models. LSIRM represents models and items in a shared space while accounting for overall model performance and item difficulty. A shorter distance between a model and an item corresponds to a higher predicted probability of a correct response, with the other estimates held fixed. We examine the resulting interaction map alongside item properties and records of correct and incorrect responses to investigate what distinguishes models with similar overall accuracy.

In BBH, nearby items on the map often share the same correct answer, such as ``Yes'' or ``No,'' or the same correct option label, such as A or B. Interpreting this grouping requires considering how answers are presented and scored. An evaluator may score the likelihood assigned to an option label, score the full option text, or extract an answer from a generated response. These procedures can produce different correctness patterns from the same model.

Previous research documents preferences for particular option labels and sensitivity to option order \citep{zheng2024}, differences between generated-answer and first-token evaluation \citep{wang2024text}, and dependence on label presentation and prediction mode \citep{nowak2026}. The effects of adjustments for response preferences also vary across evaluation conditions \citep{bhatt2026}. These findings provide reasons to investigate answer representation as an explanation for the grouping observed in LSIRM maps. In a separate evaluation across BBH, GPQA, MMLU-Pro, and MuSR, we vary answer presentation and scoring to examine how these choices affect that grouping.

To quantify the differences associated with correct-answer categories, we add terms that estimate each model's tendency to perform better or worse on items depending on which answer is correct. We estimate these terms jointly with overall model performance, item difficulty, and the remaining model--item interactions. This gives an explicit estimate of the answer-associated contribution to each model's fitted score and allows us to inspect the performance differences that remain.

In the BBH analysis, adding these terms reduces grouping by correct-answer category in the refitted map, while distinct model profiles remain. Calculating scores with and without the estimated answer contribution, while holding all fitted estimates fixed, changes model ordering. The decomposition and score comparisons let evaluators report where models differ, how those differences relate to correct-answer categories, and how their rankings depend on the contributions included in a score.

\section{A diagnostic use of LSIRM}
\label{sec:approach}

\subsection{From IRT to model--item interaction}

Item response theory (IRT) models benchmark performance through the responses of individual models to individual items. Language models are treated as respondents, benchmark questions as items, and correctness as the observed response \citep{lalor2024, vania2021}. Let $Y_{mi}\in\{0,1\}$ indicate whether model $m$ answers item $i$ correctly, and let $p_{mi}$ denote its probability of a correct response.

The Rasch, or one-parameter logistic (1PL), model separates an overall model effect from item easiness \citep{rasch1960}:
\begin{equation}
\operatorname{logit}(p_{mi})=\theta_m+\beta_i.
\label{eq:rasch}
\end{equation}
Here, $\theta_m$ represents overall model performance and $\beta_i$ represents item easiness; equivalently, $-\beta_i$ is item difficulty. A model with a larger $\theta_m$ has a higher predicted probability of success, and an item with a larger $\beta_i$ is easier for the evaluated models.

The two-parameter logistic (2PL) model adds item discrimination \citep{lord1980}:
\begin{equation}
\operatorname{logit}(p_{mi})
=
a_i(\theta_m+\beta_i),
\qquad a_i>0.
\label{eq:2pl}
\end{equation}
Discrimination allows some items to distinguish models with higher and lower overall performance more sharply than others. The term $a_i\theta_m$ allows the effect of overall model performance to vary across items, while retaining a common ordering: a model with a higher $\theta_m$ has a higher predicted probability of success on every item.

Models can also differ systematically in which items favor them. One model may perform relatively well on one group of questions and relatively poorly on another, even when its overall performance resembles that of a second model. LSIRM represents these differences by extending the Rasch model with a distance-based model--item interaction \citep{jeon2021}:
\begin{equation}
\operatorname{logit}(p_{mi})
=
\theta_m+\beta_i-\|z_m-w_i\|.
\label{eq:lsirm}
\end{equation}
The coordinates $z_m$ and $w_i$ place model $m$ and item $i$ in a shared space. Their Euclidean distance contributes to the response logit alongside the overall model and item effects. Holding those effects fixed, a shorter distance implies a higher predicted probability of a correct response. We use a two-dimensional latent space and fix the distance multiplier to one (\(\gamma=1\)). The model and item coordinates are estimated jointly with the other parameters and are not standardized to unit variance. Thus, fixing the multiplier still allows the estimated distances to vary in magnitude.

\subsection{Interpreting the interaction map}

The fitted coordinates form an interaction map. LSIRM directly models distances between models and items. The shared geometry also gives meaning to neighborhoods: nearby items have similar distance profiles to the model cohort, and nearby models have similar distance profiles to the item set. These relationships help identify groups of items that distinguish models with similar overall performance.

The map describes the interaction contribution. Predicted correctness also depends on the overall model and item effects in Equation~\ref{eq:lsirm}. We therefore examine distances alongside observed accuracy and item properties. Rotating or reflecting the map preserves its distances and fitted probabilities, so we omit coordinate axes and interpret relative positions.

\begin{figure}[t]
    \centering
\includegraphics[width=\linewidth] {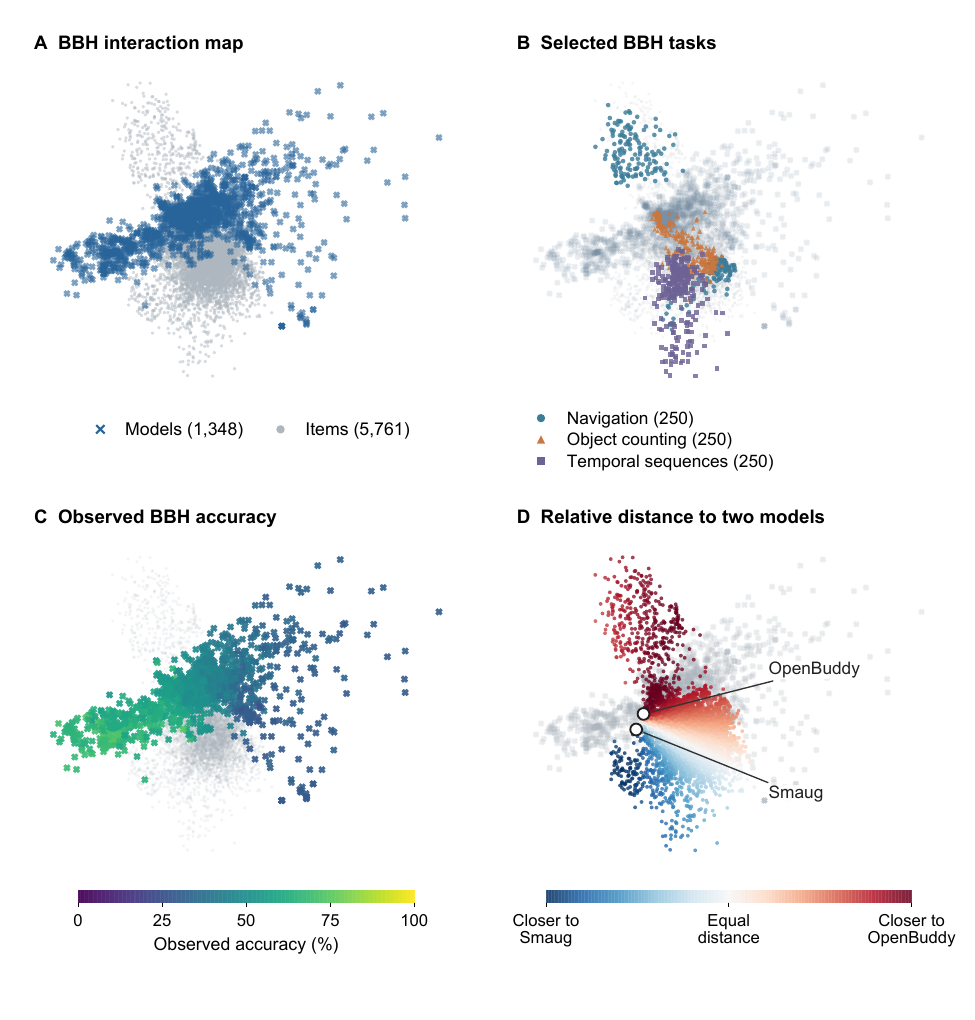}
\caption{ Interpreting a BBH interaction map through item and model annotations. All panels retain identical fitted positions, orientation, and scale. \textbf{(A)} The full map contains 1,348 models and 5,761 items. \textbf{(B)} Navigation, Object Counting, and Temporal Sequences are highlighted by task, retaining all 250 items per task. \textbf{(C)} Model points are colored by observed accuracy across the 5,761 items. \textbf{(D)} Open circles identify OpenBuddy and Smaug. Item colors represent distance to Smaug minus distance to OpenBuddy: red indicates closer proximity to OpenBuddy, blue to Smaug, and white equal distance. Color intensity reflects the size of the difference. Fitted correctness probabilities combine model--item distances with the model and item effects. }
    \label{fig:lsirm-map}
\end{figure}

Figure~\ref{fig:lsirm-map} illustrates this representation for 1,348 language models and 5,761 BBH items. Panel~A shows the full map. Panel~B highlights Navigation, Object Counting, and Temporal Sequences, retaining all 250 items from each task. These annotations connect regions of the map to the questions being evaluated and reveal variation among items within a task.

Panel~C colors models by observed BBH accuracy. Panel~D compares the distances of two similarly accurate models, OpenBuddy and Smaug, to each item. An item closer to one model has a more favorable distance contribution for that model. The comparison directs attention to items for which the two models' fitted interaction contributions differ.

\subsection{From a map pattern to a diagnostic question}

Navigation provides an example of how to investigate a pattern in the map. Its items occupy two separated concentrations. Inspection of the item records shows that one contains the 145 items whose correct answer is ``No,'' whereas the other contains the 105 items whose correct answer is ``Yes.'' Correct-answer identity therefore distinguishes the two groups within this task.

Interpreting this separation requires examining both the items and the evaluation procedure. Items with different correct answers may differ in their content or demands. Models may also perform differently depending on how answers are presented and scored. The task-by-task analysis in Section~\ref{sec:results} examines whether the association extends across BBH (Figure~\ref{fig:bbh_key_annotated}), and the protocol comparisons assess its dependence on evaluation procedures.

We then examine the consequences for model comparison. By estimating each model's tendency to perform better or worse on items with particular correct answers, we can report the associated contribution separately and inspect the interaction that remains. Scores calculated with and without this contribution show how the choice of included components affects model rankings.

\section{Data and methods}
\label{sec:methods}

We investigate the association between BBH item neighborhoods and correct answers, then estimate how that association contributes to model profiles and rankings. A separate four-benchmark evaluation examines whether changing answer presentation or scoring changes the interaction pattern. Held-out prediction assesses whether interaction improves response prediction after task and correct-answer differences are represented.

\subsection{BBH data and item neighborhoods}

The main analysis uses item-level scored responses from the Hugging Face Open LLM Leaderboard \citep{openllmv2}. The retained BBH matrix \citep{bbh2022} contains 1,348 language models and 5,761 items, coded 1 for correct and 0 for incorrect. These models form the union of 50 overlapping stratified draws of 100 models from a leaderboard-matched pool of 3,732 models. The retained data contain 24 task labels, including count-specific task variants.

All BBH decomposition and ranking comparisons use the same correctness matrix, model cohort, item order, and equal item weights. We annotate the initial LSIRM map with task labels and correct answers after fitting, then inspect the corresponding correctness records to interpret the neighborhoods. The correct-answer annotation identifies what an item requires; the response record indicates whether a model received credit.

We assess answer association within tasks because tasks differ in both content and available answers. For each item, we identify its ten nearest neighbors from the same task, excluding the item itself. In Navigation, for example, this asks whether an item whose correct answer is ``Yes'' has other ``Yes'' items among its nearest Navigation neighbors.

Let $O$ denote the average proportion of neighbors sharing the item's correct answer, and let $N$ denote the mean agreement under 200 permutations of correct-answer annotations within tasks. We calculate
\begin{equation}
A=\frac{O-N}{1-N}.
\label{eq:alignment}
\end{equation}
A value of zero corresponds to the permutation expectation; a value of one indicates that every selected neighbor shares the item's correct answer. For BBH summaries, we pool observed and permuted agreement with equal item weights before normalizing. The same permutation bank is used across fitted specifications.

\subsection{Answer presentation and scoring}

An association with correct answers may depend on how the evaluation presents and scores those answers. We examine this possibility in a separate evaluation of 18 open-weight models on 2,399 items from BBH, GPQA, MMLU-Pro, and MuSR \citep{bbh2022, gpqa2023, mmlupro2024, musr2024}. The items form 30 blocks defined by domain and number of answer options. One GPQA item was excluded across all models and conditions; its identity and recorded exclusion reason appear in Appendix~\ref{sec:collection-scope}.

We compare three protocols: text scoring without labels, text scoring with labels, and label scoring. Table~\ref{tab:protocol-example} illustrates their differences with a constructed question, ``Which animal is a mammal?'' The options and their order remain the same across conditions.

\begin{table}[t]
\centering
\small
\caption{ Constructed example of the three evaluation protocols. The question is ``Which animal is a mammal?'' The final column lists the candidate continuations whose likelihoods are compared. Parenthetical codes identify the corresponding conditions in the analysis records. }
\label{tab:protocol-example}
\begin{tabular}{p{0.25\linewidth}
                p{0.34\linewidth}
                p{0.31\linewidth}}
\toprule
Protocol & Options displayed & Candidates scored \\
\midrule
Text scoring without labels (U0)
&
\shortstack[l]{
great white shark\\
bottlenose dolphin\\
rainbow trout\\
common octopus}
&
\shortstack[l]{
great white shark\\
bottlenose dolphin\\
rainbow trout\\
common octopus}
\\[1em]

Text scoring with labels (LT0)
&
\shortstack[l]{
(A) great white shark\\
(B) bottlenose dolphin\\
(C) rainbow trout\\
(D) common octopus}
&
\shortstack[l]{
great white shark\\
bottlenose dolphin\\
rainbow trout\\
common octopus}
\\[1em]

Label scoring (L0)
&
\shortstack[l]{
(A) great white shark\\
(B) bottlenose dolphin\\
(C) rainbow trout\\
(D) common octopus}
&
\shortstack[l]{
(A)\\
(B)\\
(C)\\
(D)}
\\
\bottomrule
\end{tabular}
\end{table}

The two text-scoring protocols compare the likelihoods of the same full option strings. Their contrast measures the effect of displaying labels while keeping text scoring fixed. Text scoring with labels and label scoring use identical prompts but compare different continuations. In the example, the correct candidate is ``bottlenose dolphin'' under text scoring and ``(B)'' under label scoring.

For each protocol, the evaluator selects the candidate with the largest summed conditional log likelihood, without length normalization. The comparison between text and label scoring therefore includes differences in continuation length and tokenization.

We fit LSIRM separately to each protocol and assess alignment within each domain-by-option-count block. For the text-scoring protocols, categories identify the correct option's position; for label scoring, they identify its label. These categories correspond under the original identifier assignment. Equation~\ref{eq:alignment} is applied within each block, and the normalized values are averaged equally across blocks.

All 2,399 items enter estimation. Primary alignment contrasts summarize the 2,159 items in 27 blocks with at least three options; the display contrast across all 30 blocks is reported separately. Conditional 95\% intervals use 10,000 paired block-bootstrap draws with fitted maps held fixed.

Identifier reassignment provides a direct check of which choices models retain. We compare label scoring under the original identifiers with a condition that changes identifiers while preserving option wording and position. Across 43,182 paired choices, we record whether each model retains the same option or the same identifier. Retaining an option also retains its position in this design.

To assess cohort sensitivity, we repeat the fits with and without SmolLM2-360M-Instruct on identical items. We also fit the label-scoring data with model-specific effects for ten identifier groups to examine changes in the map and rankings after representing that association explicitly.

\subsection{Representing the answer-associated contribution}

The Navigation example suggests that correct-answer identity can distinguish items within a task. We quantify this association by allowing each model to have a different effect for each correct-answer category. For example, a model can have a positive effect for items whose correct answer is ``Yes'' and a negative effect for those whose correct answer is ``No.''

We compare three LSIRM specifications. Let $c_i$ denote item $i$'s correct-answer category and $t_i$ its task. For specification $k$, write $\eta_{mi}^{(k)}=\operatorname{logit}(p_{mi}^{(k)})$. The baseline LSIRM, M0, estimates overall model effects, item easiness, and model--item distances. M1 adds an answer effect shared across tasks. M2 allows that effect to differ by task:
\begin{subequations}
\label{eq:conditional}
\begin{align}
\text{M0:}\quad
\eta_{mi}^{(0)}
&=
\theta_m^{(0)}+\beta_i^{(0)}
-\|z_m^{(0)}-w_i^{(0)}\|,
\label{eq:m0}
\\[0.6em]
\text{M1:}\quad
\eta_{mi}^{(1)}
&=
\theta_m^{(1)}+\beta_i^{(1)}
-\|z_m^{(1)}-w_i^{(1)}\|
+\underbrace{\delta_{m,c_i}^{(1)}}_{\text{answer effect}},
\label{eq:m1}
\\[0.6em]
\text{M2:}\quad
\eta_{mi}^{(2)}
&=
\theta_m^{(2)}+\beta_i^{(2)}
-\|z_m^{(2)}-w_i^{(2)}\|
+\underbrace{\delta_{m,(t_i,c_i)}^{(2)}}_
{\text{task--answer effect}}.
\label{eq:m2}
\end{align}
\end{subequations}

In M1, each model has one effect for each correct-answer category. Its effect for ``Yes,'' for example, is shared across tasks with that category. M2 estimates a separate effect for each task--answer combination, allowing the association with ``Yes'' to vary across tasks.

M1 contains 41 answer groups and provides the primary decomposition. M2 contains 123 task--answer groups and examines whether allowing answer effects to vary by task changes the findings. Numeric Object Counting answers remain distinct in both specifications. The answer-family summaries group these 17 numeric categories under ``Other.''

For each model, the effects are centered equally across all indexed groups:
\begin{equation}
\sum_{c\in\mathcal C}\delta_{m,c}^{(1)}=0,
\qquad
\sum_{(t,c)\in\mathcal G}
\delta_{m,(t,c)}^{(2)}=0,
\label{eq:answer-centering}
\end{equation}
where $\mathcal C$ contains the 41 answer categories and $\mathcal G$ the 123 observed task--answer combinations.

The superscripts distinguish separately estimated fits. Each specification re-estimates model effects, item effects, and coordinates jointly with its answer terms. For the summaries below, we suppress the superscripts when referring to a single fit and write $I_{mi}=-\|z_m-w_i\|$ for its interaction contribution.

We examine the answer effects alongside the interaction that remains after joint estimation. For an item subset $T$, such as the 250 Navigation items, the relative interaction profile is
\begin{equation}
R_{mT}
=
\frac{1}{|T|}\sum_{i\in T}I_{mi}
-
\frac{1}{J}\sum_{i=1}^{J}I_{mi},
\label{eq:relative-interaction}
\end{equation}
where $J=5,761$. A positive value indicates a more favorable interaction contribution for that subset than for the model's average item. Observed accuracy and the estimated answer contribution accompany these profiles.

\subsection{Scores and ranking comparisons}

The decomposition allows us to compare scores that include or exclude the estimated answer contribution. Let $g(i)=c_i$ in M1 and $g(i)=(t_i,c_i)$ in M2. Define the item-weighted means
\[
\bar I_m=\frac{1}{J}\sum_{i=1}^{J}I_{mi},
\qquad
\bar\delta_m=\frac{1}{J}\sum_{i=1}^{J}
\delta_{m,g(i)}.
\]
We calculate
\begin{equation}
s_m=\theta_m+\bar I_m,
\qquad
q_m=s_m+\bar\delta_m.
\label{eq:scores}
\end{equation}
The score $s_m$ combines the overall model effect and average interaction; $q_m$ also includes the mean answer contribution. In M0, we set $\bar\delta_m=0$, so the scores coincide. Although answer effects are centered equally across groups, their item-weighted mean can differ from zero because groups contain different numbers of items.

Both scores are on the logit scale. The mean fitted logit is $q_m+\bar\beta$, where $\bar\beta$ is the mean item easiness effect. Within a fit, every model is evaluated on the same items, so this common item term does not affect the ranking. Mean predicted accuracy would instead average the fitted probabilities across items.

Within-fit comparisons hold every estimate fixed and rank models by $q_m$ and $s_m$. Excluding the answer contribution changes each score by exactly $-\bar\delta_m$: a positive contribution lowers the score when removed, whereas a negative contribution raises it. Comparing the resulting rankings shows how ordering depends on including that contribution.

Between-fit comparisons rank models by $s_m$ from M0, M1, and M2. Here, adding answer terms leads to re-estimation of model effects, item effects, and coordinates. Score differences therefore include changes in the overall model effects and average interactions. The additive decomposition applies within each fitted specification.

We rank unrounded scores in descending order and assign average ranks to exact ties. Every BBH comparison retains the same 1,348 models and 5,761 equally weighted items. Full-cohort summaries describe the distribution of rank changes, while selected examples show contrasting outcomes.

For the protocol evaluation, we calculate $s_m$ over all 2,399 items. Rankings from label scoring, with and without identifier effects, are compared with text scoring without labels as an alternative protocol reference. We also re-rank the shared 17 models within the 18-model fits, separating changes among those models from insertion of the additional model. The intended use of an evaluation determines which protocol and score components should define its ranking.

\subsection{Prediction and numerical stability}

Held-out prediction assesses whether interaction captures response variation after differences across tasks and correct answers have been represented. We use one fixed 100-model BBH draw, selected among the 50 draws for performance and metadata representativeness. A model-by-task--answer specification is compared with its two-dimensional interaction extension using paired held-out log loss and Brier score.

Each of ten overlapping masks withholds 57,400 responses and retains 518,700 for estimation. The masking pool contains 5,740 items with at least two observed successes and two failures; the other 21 items remain in training. All 5,761 items enter each fit, and held-out responses enter neither estimation nor initialization. Both specifications use the historical 107-group coding, which combines numeric object-counting answers into one task-specific group. The comparison evaluates response prediction within this fixed model cohort and item pool.

We fit all specifications by minimizing the summed Bernoulli negative log likelihood with quadratic penalties using L-BFGS-B. The spread of the estimated coordinates is determined by the response likelihood and the fixed quadratic penalties specified in the supplementary methods. No separate coordinate-variance parameter is estimated.
Numerical acceptance requires first-order stationarity and agreement between independently computed objectives and gradients. Additional random starts and local perturbations assess the sensitivity of the BBH decomposition, alignment, and rankings. Where multiple accepted solutions are available, we retain the one with the lowest original penalized objective. Penalty coefficients, initialization procedures, numerical tolerances, and starting-value comparisons appear in Appendix~\ref{sec:numerical}.

\section{Results}
\label{sec:results}

\subsection{Correct-answer categories distinguish item neighborhoods within tasks}

\begin{figure}[p]
    \centering
\includegraphics[ width=\linewidth, height=0.82\textheight, keepaspectratio ]{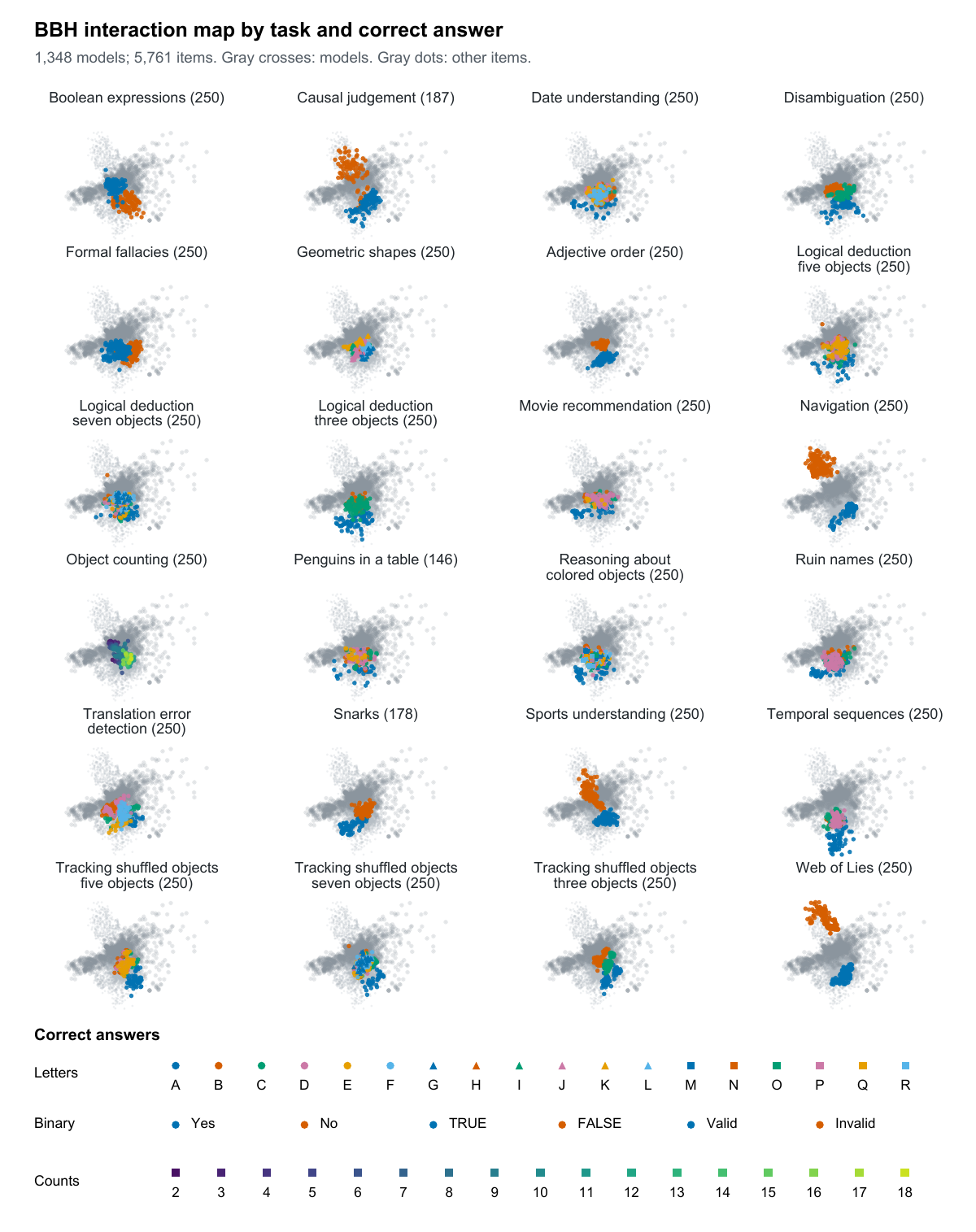}
\caption{ BBH interaction map annotated by task and correct answer. The map was estimated from the correctness records of 1,348 models on 5,761 items. Each panel highlights the items from one task, with colors and marker shapes indicating their correct answers. Gray crosses represent models; gray dots represent items from other tasks. Panel titles report task item counts. All panels use the same fitted positions, orientation, and scale. Correct-answer categories were added as annotations after fitting. Coordinate axes are omitted because their orientation has no substantive interpretation. }
    \label{fig:bbh_key_annotated}
\end{figure}

Items from the same BBH task can occupy different regions of the interaction map depending on their correct answers (Figure~\ref{fig:bbh_key_annotated}). This separation is especially clear in Navigation and Web of Lies, where items with ``Yes'' as the correct answer form a different neighborhood from those with ``No'' as the correct answer. Other tasks, including Date Understanding and Logical Deduction with seven objects, show more overlap between answer categories. The task-by-task view therefore reveals both the answer-associated pattern and differences in its strength across tasks.

The alignment measure in Equation~\ref{eq:alignment} summarizes how strongly correct-answer categories distinguish item positions within tasks. In the baseline fit, alignment is highest for Yes/No items ($0.9629$), followed by True/False ($0.8935$), Valid/Invalid ($0.7009$), and letter answers ($0.3650$). These differences motivate a closer examination of the response patterns behind the map.

\begin{figure}[p]
\centering
\includegraphics[width=\linewidth]{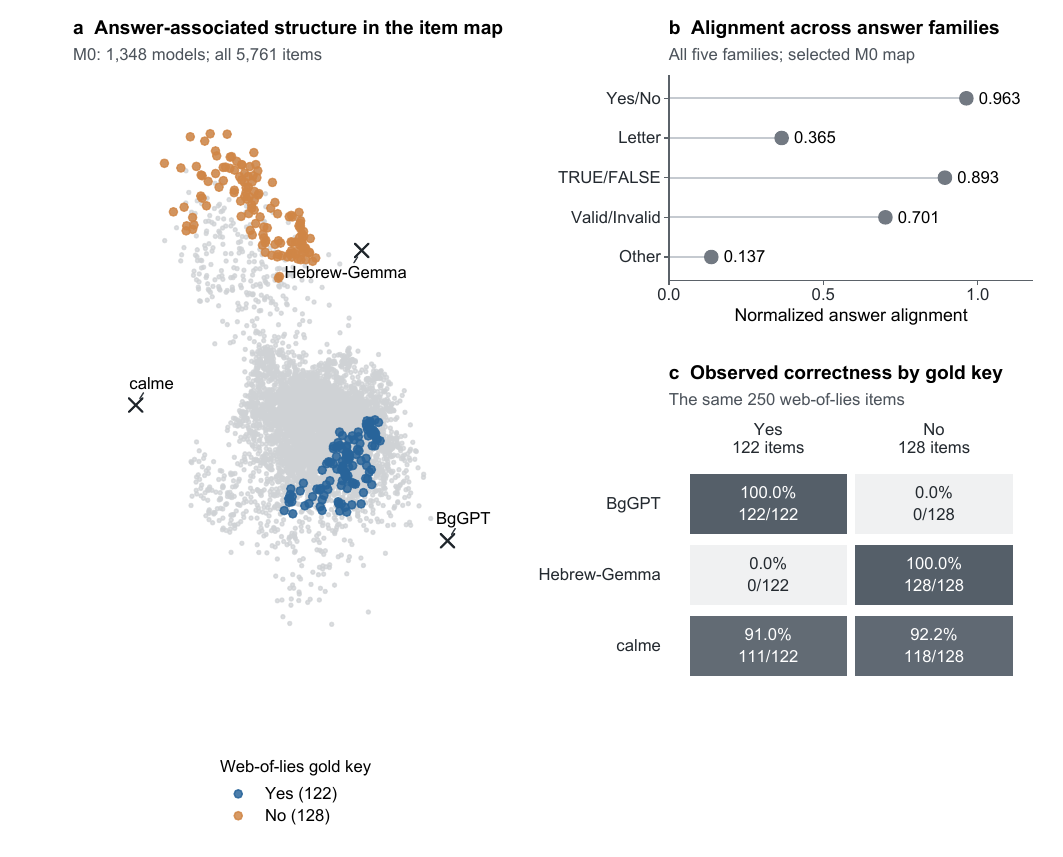}
\caption[Answer-associated structure and observed response profiles] {\input{notes/figure1_caption}}
\label{fig:diagnosis}
\end{figure}

Web of Lies provides a concrete example. Its two item neighborhoods correspond to sharply different performance profiles among individual models (Figure~\ref{fig:diagnosis}). BgGPT answers all 122 items whose correct answer is ``Yes'' correctly and none of the 128 whose correct answer is ``No.'' Hebrew-Gemma shows the opposite pattern, answering none of the ``Yes'' items and all 128 ``No'' items correctly. The calme model answers most items correctly in both groups: 111 of 122 ``Yes'' items and 118 of 128 ``No'' items. These correctness profiles connect the item neighborhoods to differences among models; full model identifiers appear in Table~\ref{tab:s-example-identities}.

\subsection{Scoring labels increases answer alignment at a fixed prompt}

\begin{figure}[p]
\centering
\includegraphics[width=\linewidth]{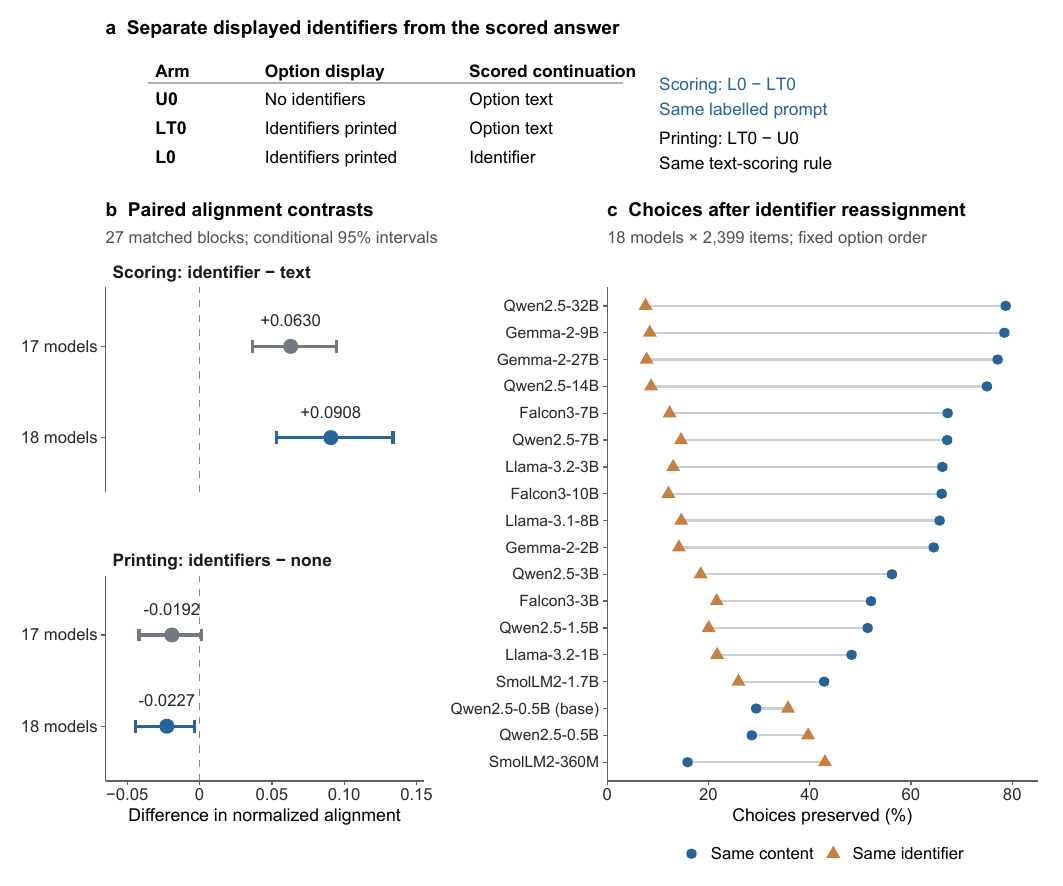}
\caption[Effects of answer display and scoring] {\input{notes/figure2_caption}}
\label{fig:protocol}
\end{figure}

The four-benchmark evaluation shows that the scored answer representation affects alignment even when the prompt is unchanged (Figure~\ref{fig:protocol}). Across the 27 matched blocks containing 2,159 items, label scoring produces greater alignment than text scoring with labels. The increase is 0.0630 in the 17-model cohort, with a conditional 95\% interval of [0.0365, 0.0945]. Including the eighteenth model gives an increase of 0.0908 [0.0534, 0.1338] (Table~\ref{tab:s-crossed-alignment}). These intervals describe variation across blocks with the fitted maps held fixed.

Displaying labels while continuing to score option text has a smaller, negative estimated contrast: $-0.0192$ [$-0.0419$, 0.0011] for 17 models and $-0.0227$ [$-0.0446$, $-0.0036$] for 18 models. The increase in alignment thus occurs when the evaluator switches from scoring option text to scoring labels under the same labelled prompt. Under the raw likelihood rule used here, that switch also changes the length and tokenization of the scored continuation.

The choice records distinguish this map pattern from persistence in selecting a particular label. After labels are reassigned, models retain the same option in 57.27\% of the 43,182 paired choices and the same label in 18.86\%. Retaining the option is more common for 15 of the 18 models. Option wording and position remain fixed in this comparison, so retaining the option also retains its position. Greater alignment under label scoring therefore coexists with more frequent preservation of the selected option than of its label.

This dependence on answer representation is consistent with previous findings on option-label preferences and evaluation procedures \citep{zheng2024,wang2024text,nowak2026}. The present comparison shows how that dependence appears in an LSIRM map under the specified likelihood scoring rules. Together with the BBH response profiles, it motivates estimating the contribution associated with correct-answer categories explicitly.

\subsection{The decomposition distinguishes models with similar accuracy}

\begin{figure}[p]
\centering
\includegraphics[width=.98\linewidth] {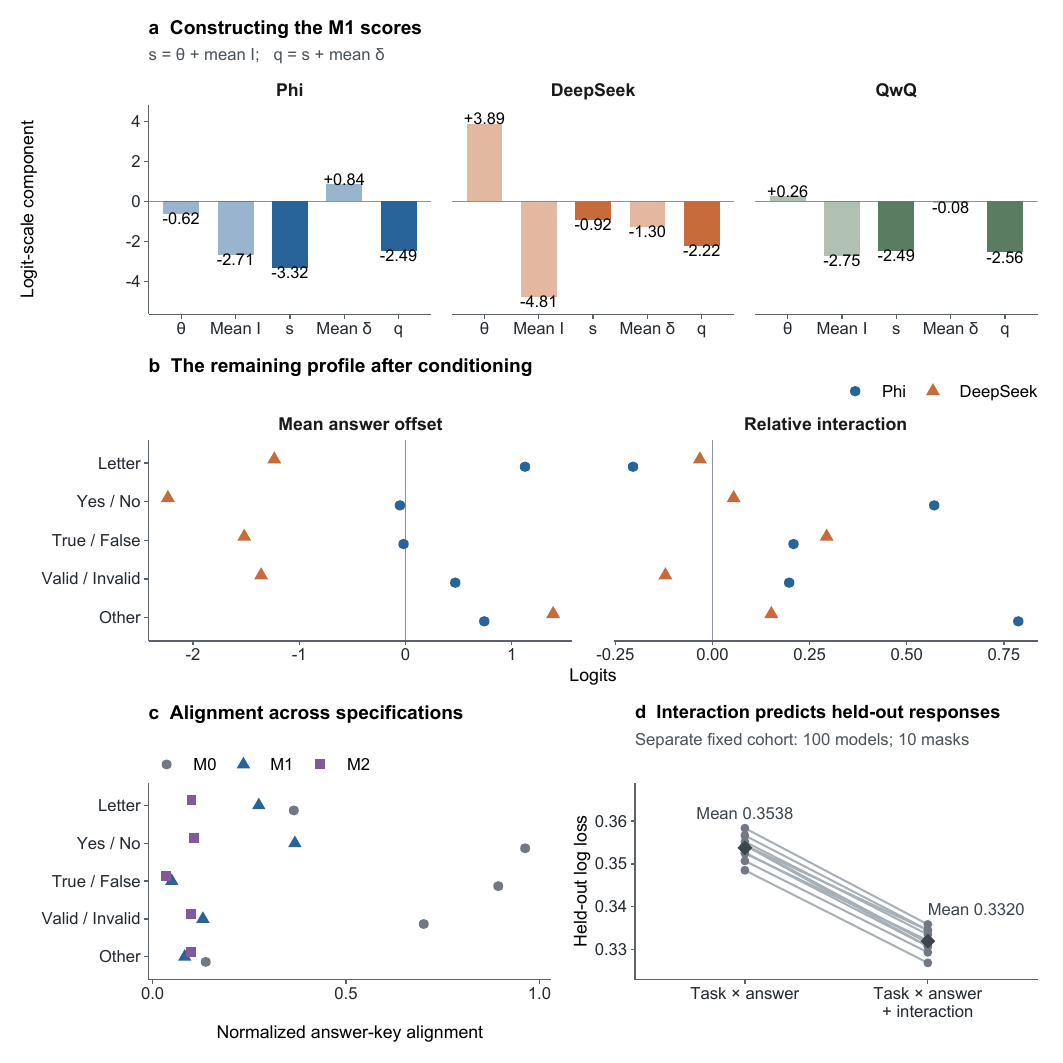}
\caption[Answer contributions and remaining performance profiles] {\input{notes/figure3_caption}}
\label{fig:components}
\end{figure}

Adding answer effects reduces the association between correct answers and the remaining item geometry (Figure~\ref{fig:components}). On the same BBH response matrix, Yes/No alignment falls from 0.9629 in M0 to 0.3677 with pooled answer effects in M1 and 0.1077 with task--answer effects in M2. True/False alignment falls from 0.8935 to 0.0496 and 0.0352, respectively. All five answer families have lower alignment under both specifications than under M0 (Table~\ref{tab:s-family-alignment}). These reductions occur after joint re-estimation of the answer terms, model and item effects, and interaction coordinates.

Phi and DeepSeek illustrate how this decomposition distinguishes models with similar overall accuracy. They answer 3,253 and 3,239 items correctly, respectively, a difference of 14 out of 5,761 items. Yet on 2,158 items, exactly one of the two models is correct. Their similar totals therefore combine many opposing successes and failures.

In M1, the estimated mean answer contribution is $+0.838$ logits for Phi and $-1.302$ for DeepSeek. The contribution raises Phi's answer-included score $q$ relative to its answer-excluded score $s$ and lowers DeepSeek's. Figure~\ref{fig:components}a and Table~\ref{tab:s-components} show how the overall model effect, mean interaction, and mean answer effect combine for each model. All components are calculated over the same 5,761 items and add on the fitted logit scale.

The remaining interaction profiles locate differences within each model's performance pattern. Among Yes/No items, Phi has a relative interaction of $+0.571$ and a mean answer effect of $-0.053$; DeepSeek's corresponding values are $+0.055$ and $-2.237$. Relative interaction compares an item group with that model's own all-item mean. It therefore describes where the distance contribution is more or less favorable within the model's profile.

Object Counting shows why this relative profile should be read alongside observed accuracy. Phi has a larger relative interaction than DeepSeek, 0.787 versus 0.151, while its observed accuracy is lower, 39.2\% versus 61.2\%. The mean answer contributions are positive for both models, at 0.741 and 1.389. Fitted probabilities combine these terms with the overall model and item effects. The complete task profiles also retain the contrasts in adjective order, or hyperbaton (92.0\% versus 56.0\% accuracy), and geometric shapes (21.6\% versus 55.6\%; Figure~\ref{fig:task-profiles}).

Interaction also contributes to held-out prediction after task--answer effects are included. In the separate 100-model cohort with 107 task--answer groups, adding two-dimensional interaction lowers mean held-out log loss from 0.3538 to 0.3320, with improvement in every one of the ten overlapping masks. Mean Brier score falls from 0.1118 to 0.1047. Mask-level log losses appear in Table~\ref{tab:s-heldout}. The gain concerns responses held out from the same model cohort and item pool.

\subsection{Including the answer contribution changes model rankings}

\begin{figure}[p]
\centering
\includegraphics[width=\linewidth] {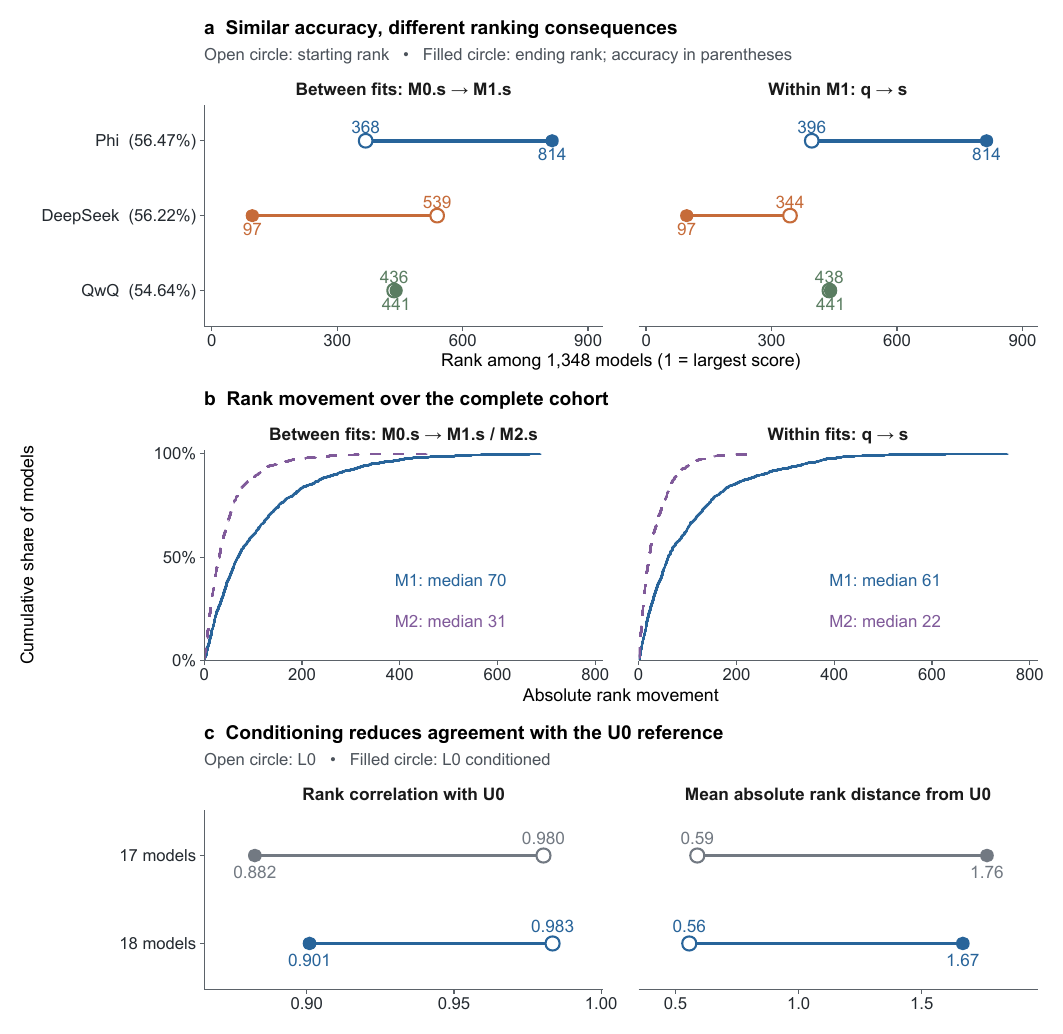}
\caption[Ranking changes within and between fitted models] {\input{notes/figure4_caption}}
\label{fig:rankings}
\end{figure}

The opposite answer contributions for Phi and DeepSeek produce opposite ranking movements within M1 (Figure~\ref{fig:rankings}). With all estimates held fixed, excluding the answer contribution lowers Phi's score by 0.838 logits and raises DeepSeek's by 1.302 logits. Re-ranking the same 1,348 models moves Phi from 396 on $q$ to 814 on $s$, and DeepSeek from 344 to 97. Their answer-excluded ranks are 717 places apart.

Each score change follows directly from the model's mean answer contribution. Rank movement also depends on how the other models' scores change. QwQ provides a low-movement example, selected by the rule in Appendix~\ref{sec:example-selection}: excluding its answer contribution moves its rank from 438 to 441. Across the full cohort, the median absolute within-fit movement is 61 places for M1 and 22 for M2 (Figure~\ref{fig:rankings}b).

Refitting the model produces a separate ranking comparison. Between M0 and M1, Phi moves from 368 to 814 on $s$, DeepSeek from 539 to 97, and QwQ from 436 to 441. These changes include re-estimation of the overall model effects and interaction after answer terms are added.

Across all models, the median absolute movement from M0.$s$ is 70 places for M1.$s$ and 31 for M2.$s$. The corresponding rank correlations are 0.9190 and 0.9842. Of the models in M0's top 50, 28 remain in M1's top 50 and 40 in M2's (Table~\ref{tab:s-rank-comparisons}). Both the fitted specification and the decision to include its answer contribution therefore affect model ordering.

The main ranking contrasts persist across the saved local solutions. Among the preceding best M1 fit and four reoptimized perturbations, Phi's $s$ rank ranges from 814 to 816 and DeepSeek's remains at 97. All five answer-family alignment summaries remain below M0 across the saved candidates for both M1 and M2. Detailed profiles show more variation; Tables~\ref{tab:s-start-stability} and \ref{tab:s-example-start-ranges} report the component and ranking ranges.

The protocol evaluation shows a further consequence of changing the fitted specification. Adding label effects and comparing the resulting $s$ rankings reduces agreement with text scoring without labels. In the 17-model cohort, Spearman correlation with that reference falls from 0.9804 to 0.8824, while mean absolute rank distance rises from 0.588 to 1.765 places. The corresponding 18-model values are 0.9835 to 0.9009 and 0.556 to 1.667 (Figure~\ref{fig:rankings}c; Table~\ref{tab:s-crossed-ranks}).

Including the eighteenth model leaves the ordering of the shared 17 models unchanged in the selected fits; SmolLM2-360M-Instruct ranks eighteenth under each compared specification. In this evaluation, fitting label effects and ranking models by the resulting answer-excluded scores does not recover the ordering obtained by scoring option text. The ranking consequence of representing an answer effect is therefore distinct from the consequence of changing the scoring protocol.

\section{Discussion}

LSIRM helps evaluators examine differences in benchmark performance that overall accuracy conceals. By accounting for overall model performance and item difficulty, it identifies patterns in which models succeed on which items. This study demonstrates how to interpret those patterns using item properties and response records, investigate the role of evaluation procedures, and quantify the contribution associated with a particular item property. The BBH example connects this diagnostic process to concrete outcomes: model profiles that distinguish similarly accurate models and score comparisons that show how including or excluding an estimated contribution changes their rankings.

The task-by-task maps reveal that correct-answer categories distinguish item neighborhoods within several BBH tasks. Previous research provides reasons to investigate this pattern. Studies have documented preferences for option identifiers, differences between generated-answer and first-token evaluation, and sensitivity to label presentation and prediction mode \citep{zheng2024,wang2024text,nowak2026}. Our comparison across BBH, GPQA, MMLU-Pro, and MuSR examines how these evaluation choices affect the interaction maps. With the labelled prompt held fixed, scoring option identifiers produces stronger alignment with correct-answer categories than scoring option text. Agreement between the map, previous findings, and this controlled comparison supports interpreting part of the observed structure in relation to answer representation and scoring.

The decomposition makes the association with correct answers explicit for individual models. Phi and DeepSeek differ by only 14 correct responses out of 5,761 items, yet their correctness records differ on more than a third of those items. Their estimated mean answer contributions have opposite signs, and their remaining interaction profiles differ across tasks and answer families. These outputs help explain how models with similar totals can have different patterns of performance. Reading the fitted contributions alongside observed task accuracy shows where an answer-associated pattern accompanies a performance difference and where further differences remain. Because the model effects, item effects, answer effects, and interactions are estimated jointly, the decomposition depends on the chosen specification. An answer effect can combine associations with answer content, option labels, and option positions.

The ranking comparisons show the consequences of deciding which fitted contributions enter a score. Within one fit, comparing the score that includes the answer contribution ($q$) with the score that excludes it ($s$) holds every parameter estimate fixed. The resulting rank changes isolate the effect of that scoring decision. Comparing scores across separately fitted specifications also reflects changes in the estimated model effects and interactions. In the crossed evaluation, this between-fit adjustment reduced agreement with the ranking obtained by scoring option text without labels in both the 17- and 18-model cohorts. Fitting answer effects and ranking models by the resulting answer-excluded scores did not recover that reference ordering. A preferred ranking requires a stated evaluation target, such as performance under the protocol intended for deployment. The dependence of response-bias calibration on evaluation conditions provides a further reason to assess any adjustment against that target \citep{bhatt2026}.

The remaining interaction also contains information useful for prediction. In the separate 100-model analysis, including interaction improved prediction for held-out responses after task--answer effects were included. This result supports retaining interaction when modeling responses within that cohort; interpreting particular neighborhoods requires evidence about their items and response patterns. The numerical checks address a different question. The selected fits met the stationarity criteria, and alignment and rankings remained similar across the examined local solutions, although some detailed components varied. Broad comparisons were more stable than every detail of the fitted profiles.

The analyses concern the models, items, and protocols represented in the data. The main BBH cohort combines overlapping draws, and the prediction analysis reuses one selected cohort under overlapping response masks. These designs support comparisons within the analyzed cohorts rather than population-wide estimates for language models. Incomplete version and collection metadata also limit reconstruction of the original benchmark evaluations.

For practical evaluation, an LSIRM report can present observed task accuracy, the estimated answer contribution, and the remaining interaction profile together. Evaluators can use the task profiles to examine performance relevant to an intended application and trace unusual patterns back to individual response records. Scores calculated with and without the answer contribution should use the same models, items, and item weights, so their comparison has a clear interpretation. Reporting both rankings makes the scoring decision visible and allows evaluators to judge which contributions belong in a comparison for their intended use.

\paragraph{Data and code availability.}
The supplementary material provides model identifiers, component and ranking summaries, and numerical checks. Analysis code, processed response matrices, and machine-readable source tables prepared for this study are not currently publicly available.

\FloatBarrier
\begingroup\raggedright

\endgroup
\clearpage
\appendix
\renewcommand{\thefigure}{S\arabic{figure}}
\renewcommand{\thetable}{S\arabic{table}}
\renewcommand{\theHfigure}{supp.\arabic{figure}}
\renewcommand{\theHtable}{supp.\arabic{table}}
\setcounter{figure}{0}
\setcounter{table}{0}
\section{Supplementary methods and provenance}

\subsection{Estimation and numerical checks}
\label{sec:numerical}
All retained models use the summed Bernoulli negative log likelihood with penalty
\begin{equation}
0.05\|\theta\|^2+0.005\|\beta\|^2+
0.2222(\|Z\|_F^2+\|W\|_F^2)+0.02\|\delta\|_F^2,
\end{equation}
omitting absent terms. Here, $Z$ and $W$ collect the model and item coordinates, respectively, and $\|\cdot\|_F$ denotes the Frobenius norm. The implemented interaction is $-\sqrt{\max(d^2,0)+10^{-8}}$, where $d^2$ is the expanded squared Euclidean distance. Geometry has two dimensions and its multiplier is fixed to one. For each model, offsets are centered equally over all indexed groups: 41 in BBH M1, 123 in BBH M2, 107 in the prediction comparison, and ten in crossed L0 conditioning. Task--answer coding does not imply separate centering within each task.

Initial SVD fits use marginal proportions clipped to [0.01, 0.99], additive logits, and residual SVD coordinates with square-root singular-value scaling. Held-out cells enter neither marginal calculations nor initialization. The BBH numerical sequence includes the initial multistart fits, additional random starts, and perturbations of the preceding best M1 and M2 solutions. Four local perturbations per conditioned specification were accepted. Larger perturbations found lower objectives, and the selected display solutions are those lower accepted endpoints. They were not themselves subjected to a further perturbation cycle. The local sensitivity summaries compare each preceding anchor with its four perturbation endpoints, giving five candidates per specification.

Acceptance requires original and independent maximum absolute gradients at most $10^{-4}$, objective agreement within $10^{-6}$, gradient agreement within $10^{-7}$, centered-offset row sums within $10^{-10}$, finite values, and a nonincreasing objective within $10^{-6}$. The selected M0, M1, and M2 gradients are all below $10^{-5}$ (Table~\ref{tab:s-numerical-acceptance}). All eight local perturbation endpoints meet the criteria, although recovery of the preceding objective differs by perturbation size. These checks establish stationarity of the retained endpoints under the stated objective. They do not establish a global optimum or a unique decomposition.

Across the five local candidates, the maximum difference in a family-alignment summary is 0.009063 for M1 and 0.000998 for M2. Phi's M1.$s$ rank ranges from 814 to 816, DeepSeek's stays at 97, and QwQ's ranges from 441 to 444. Their M2.$s$ ranks remain 615, 229, and 434. Finer profiles can be less stable. For the 25 object-counting items with correct answer 8, the M1 relative-interaction range is 0.151 logits for Phi and 0.557 for DeepSeek. These are descriptive starting-solution ranges, not confidence intervals.

The prediction analysis retains 20 accepted fits, two specifications for each of ten masks. Two endpoints required tighter continuations to resolve independent numerical-agreement failures; the original record had accepted 18 of 20. The crossed check contains 16 accepted fits: four specifications in two cohorts, each using an SVD start and a start transferred from the other cohort. Reciprocal starts are related candidates, not independent random replicates. Selection always uses the original penalized objective, never alignment, rank movement, or held-out performance. The recorded fitting environment is Python 3.12.9, NumPy 2.5.4, SciPy 1.18.1, and JAX 0.12.0.1.

\subsection{Item weighting, prediction masks, and crossed sensitivity}
BBH alignment uses ten nearest item neighbors within each task, excluding the focal item. Canonical item order resolves distance ties. We use a shared bank of 200 within-task permutations (seed 20261025), pool observed and null agreement with equal item weights within a reported family, and then apply Equation~\ref{eq:alignment}. The ``Other'' family contains Object Counting items and retains their 17 distinct numeric answer keys. For crossed maps, the same calculation is performed within each domain-by-option-count block using 200 shared within-block permutations (seed 20261026). Normalized block values are then averaged equally.

The crossed conditional intervals use 10,000 paired block-bootstrap draws. The same sampled blocks enter both arms of each contrast; fitted coordinates remain fixed. Seeds 20261027 and 20261028 generate the matched-27-block and full-30-block resampling banks, respectively; each bank is shared across arms and cohorts. The primary matched set has 27 blocks and 2,159 items. The full printing contrast includes 30 blocks and 2,399 items (Table~\ref{tab:s-crossed-alignment}). Scoring contrasts remain positive across the retained starting solutions. The selected increase in the scoring contrast after adding the eighteenth model is 0.0278, with conditional interval [0.0052, 0.0546]. Across all four starting-solution pairings the increase ranges from 0.0168 to 0.0295, and one interval includes zero. Full-30-block alignment is more start-sensitive than the primary matched-block summaries. Crossed rank comparisons vary by at most one place across retained starts; within-specification rank correlations are at least 0.99755 and top-five membership is unchanged.

Each prediction mask withholds 57,400 cells from the fixed $100\times5,761$ matrix. Eligibility requires at least two observed successes and two failures per item, yielding 5,740 eligible items; the other 21 stay wholly in training. Mask 4 covers 5,739 distinct held-out items. Across ten masks there are 574,000 held-out occurrences but 373,751 distinct cells. Training-only estimation leaves 518,700 cells per mask. The historical 107-group coding collapses the 17 numeric object-counting keys into one task-specific group. Mask-level losses are reported in Table~\ref{tab:s-heldout} without treating masks as independent populations.

\subsection{Example selection and response verification}
\label{sec:example-selection}
Five previously named models were retained for verification: BgGPT, Hebrew-Gemma, calme, Phi, and DeepSeek. A sixth, QwQ, was selected to supply a concrete low-movement comparison. Eligible models had M0.$s$ ranks within the central 80\% of the cohort and observed accuracy between 0.2 and 0.7, excluding the five named models. Among the 1,076 eligible models, selection minimized the largest absolute movement across M0.$s$--M1.$s$, M0.$s$--M2.$s$, M1.$q$--M1.$s$, and M2.$q$--M2.$s$, breaking ties by total movement and then model identifier. This is a documented illustrative selection, not a population-sampling rule.

All 34,566 correctness cells for the six models were matched to their source item/model/score records, with no selected duplicate records. Component summaries were independently recomputed from the saved coordinates and parameters. For Phi and DeepSeek, 2,167 items are correct for both, 1,436 incorrect for both, 1,086 correct only for Phi, and 1,072 correct only for DeepSeek. Full task profiles and full identifiers accompany the manuscript. No chosen-answer string is inferred from a correctness-only record.

Four additional models provide the introductory comparisons (Figure~\ref{fig:introduction-comparisons}). OpenBuddy and Smaug illustrate a reversal across tasks among two 70B models with similar BBH accuracy. The Mistral 7B Base-SFT-CPO and Base-SFT-RDPO pair illustrates a reversal across correct-answer categories within Ruin Names. These pairs were selected from the same response matrix to illustrate contrasting observed profiles. Each pair's 11,522 correctness entries was separately matched to the source records, with no duplicate retained records. Table~\ref{tab:s-example-identities} lists all ten example identities. Navigation and Adjective order are the display names for the archived tasks \texttt{navigate} and \texttt{hyperbaton}; these aliases retain the same item sets.

\subsection{Collection records and analysis scope}
\label{sec:collection-scope}
The complete crossed table contains 499,482 observations across 16 arms. The main set consists of 2,399 items in 30 blocks; a separate binary-answer context set has 480 items in six blocks. The common excluded item is \texttt{gpqa:gpqa\_extended\_356}. Collection code attributes this exclusion to question-stem truncation; the underlying diagnostic log was not recovered. The historical SmolLM2-360M-Instruct exclusion appears in the fitting code and archived screening table, but the screen-generating code and threshold were not recovered. The current analysis directly checks inclusion rather than assigning a behavioral interpretation to its near-chance score.

The retained direct-choice comparison uses the first label reassignment, in which option wording and order are fixed and identifiers change by item-specific nonidentity cyclic offsets. Content preservation and identifier preservation are disjoint in this comparison; the remaining 23.87\% preserves neither. The second reassignment excludes each item's first offset and is not an independent draw. Reduced-option arms jointly change distractors and correct-option placement. These ancillary arms are not used to establish the present decomposition or ranking results.

The observational files identify models, items, and correctness, with exact hashes and ordered manifests. All 1,348 identifiers join to the saved leaderboard; 1,345 have recorded commit hashes. These hashes are leaderboard metadata, not verified bindings to the response runs. The original retrieval date, response-run model revisions, and complete observational harness metadata were not recovered. Crossed collection scripts request Python 3.11, lm-eval 0.4.13, and transformers 5.17.0, with bfloat16 for Gemma models and float16 otherwise. Complete executed version records and resolved model revisions are unavailable, so requested settings are distinguished from verified runtime metadata.

The available crossed raw archive contains 479,411 retained records, all matching the canonical merged table. Nine Gemma-27B arm files, totaling 20,071 retained observations, were not recovered locally; their records are present in the complete merged table used for fitting. No crossed per-option likelihood archive or completed length-normalized outcome analysis was recovered. The implemented response reducer uses the argmax of raw summed continuation likelihoods. These records support the reported correctness analyses while leaving parts of the original collection provenance incomplete.

The archived project account identifies choice preservation as its primary crossed comparison and the scoring contrast as secondary; it states that the plan was not independently registered. An independently dated original plan was not recovered. The present fit refinements and 17-versus-18-model check are subsequent analyses of existing responses. The local source ledger preserves their lineage and replaces earlier numerical summaries with the verified endpoints used here.

\clearpage
\section{Supporting results}

\begin{table}[htbp]
\centering\small
\caption{Rank comparisons for all 1,348 BBH models on the same 5,761 items. Lower ranks indicate larger scores. Between-fit comparisons re-estimate all components; within-fit comparisons remove only the mean answer offset. The last column counts models shared by the two top-50 sets.}
\label{tab:s-rank-comparisons}
\begin{tabular}{@{}llrrrr@{}}
\toprule
Comparison & Scores & $\rho$ & Median move & Max. move & Top-50 overlap \\
\midrule
between fit & M0.s$\to$M1.s & 0.9190 & 70 & 687 & 28 \\
between fit & M0.s$\to$M2.s & 0.9842 & 31 & 454 & 40 \\
within fit & M1.q$\to$M1.s & 0.9297 & 61 & 755 & 30 \\
within fit & M2.q$\to$M2.s & 0.9920 & 22 & 239 & 43 \\
\bottomrule
\end{tabular}
\end{table}

\begin{table}[htbp]
\centering\small
\caption{Normalized answer-family alignment on the common BBH item set. Neighborhoods contain ten items within the same task; the null uses 200 within-task permutations. Other contains all 250 Object Counting items, with their 17 numeric answer keys retained separately. Values are selected-fit summaries.}
\label{tab:s-family-alignment}
\begin{tabular}{@{}lrrrr@{}}
\toprule
Correct-answer family & Items & M0 & M1 & M2 \\
\midrule
Letter & 4074 & 0.3650 & 0.2745 & 0.1006 \\
TRUE/FALSE & 250 & 0.8935 & 0.0496 & 0.0352 \\
Yes/No & 937 & 0.9629 & 0.3677 & 0.1077 \\
Valid/Invalid & 250 & 0.7009 & 0.1300 & 0.0980 \\
Other & 250 & 0.1373 & 0.0830 & 0.0984 \\
\bottomrule
\end{tabular}
\end{table}

\begin{table}[htbp]
\centering\small
\caption{Exact identities of the ten BBH examples used in the text and figures. Correctness is calculated over the same 5,761 items per model. Accuracy is a proportion. The introductory pairs and the six diagnostic/ranking examples have separate source-record checks described in Appendix~\ref{sec:example-selection}.}
\label{tab:s-example-identities}
\begin{tabularx}{\linewidth}{@{}lXrr@{}}
\toprule
Short name & Model identifier & Correct & Accuracy \\
\midrule
OpenBuddy & \nolinkurl{OpenBuddy/openbuddy-llama3-70b-v21.2-32k} & 3743 & 0.6497 \\
Smaug & \nolinkurl{abacusai/Smaug-Llama-3-70B-Instruct-32K} & 3736 & 0.6485 \\
CPO & \nolinkurl{princeton-nlp/Mistral-7B-Base-SFT-CPO} & 2514 & 0.4364 \\
RDPO & \nolinkurl{princeton-nlp/Mistral-7B-Base-SFT-RDPO} & 2531 & 0.4393 \\
BgGPT & \nolinkurl{INSAIT-Institute/BgGPT-Gemma-2-27B-IT-v1.0} & 1657 & 0.2876 \\
Hebrew-Gemma & \nolinkurl{yam-peleg/Hebrew-Gemma-11B-Instruct} & 2314 & 0.4017 \\
calme & \nolinkurl{MaziyarPanahi/calme-2.4-rys-78b} & 4182 & 0.7259 \\
Phi & \nolinkurl{Syed-Hasan-8503/Phi-3-mini-4K-instruct-cpo-simpo} & 3253 & 0.5647 \\
DeepSeek & \nolinkurl{deepseek-ai/DeepSeek-R1-Distill-Llama-70B} & 3239 & 0.5622 \\
QwQ & \nolinkurl{prithivMLmods/QwQ-LCoT1-Merged} & 3148 & 0.5464 \\
\bottomrule
\end{tabularx}
\end{table}

\begin{table}[htbp]
\centering\small
\caption{M1 score construction for the ranking examples. All components are on the logit scale and use equal weights over 5,761 items: $s=\theta+\bar I$ and $q=s+\bar\delta$. The model-independent mean item effect is omitted from both scores. Components and scores are rounded after calculation. Values are conditional on the selected numerically accepted fit.}
\label{tab:s-components}
\begin{tabular}{@{}lrrrrrrr@{}}
\toprule
Model & $\theta$ & $\bar I$ & $\bar\delta$ & $s$ & $q$ & Rank $s$ & Rank $q$ \\
\midrule
Phi & -0.616 & -2.708 & 0.838 & -3.323 & -2.486 & 814 & 396 \\
DeepSeek & 3.886 & -4.806 & -1.302 & -0.920 & -2.222 & 97 & 344 \\
QwQ & 0.262 & -2.748 & -0.076 & -2.486 & -2.561 & 441 & 438 \\
\bottomrule
\end{tabular}
\end{table}

\begin{table}[htbp]
\centering\small
\caption{Paired held-out log loss for one fixed 100-model BBH cohort and ten overlapping response masks. Each mask withholds 57,400 responses and retains 518,700 for estimation. Both specifications use the historical 107 task--answer groups. Gain is the task--answer-only loss minus the loss after adding interaction; it is not model-population uncertainty.}
\label{tab:s-heldout}
\begin{tabular}{@{}rrrr@{}}
\toprule
Mask & Task--answer only & With interaction & Gain \\
\midrule
1 & 0.354357 & 0.330746 & 0.023611 \\
2 & 0.350713 & 0.329354 & 0.021359 \\
3 & 0.352481 & 0.332135 & 0.020346 \\
4 & 0.352671 & 0.330756 & 0.021915 \\
5 & 0.354487 & 0.331522 & 0.022965 \\
6 & 0.348527 & 0.326901 & 0.021626 \\
7 & 0.355191 & 0.333507 & 0.021684 \\
8 & 0.358360 & 0.335874 & 0.022486 \\
9 & 0.354385 & 0.334492 & 0.019893 \\
10 & 0.356642 & 0.334561 & 0.022081 \\
Mean & 0.353781 & 0.331985 & 0.021797 \\
\bottomrule
\end{tabular}
\end{table}

\begin{table}[htbp]
\centering\small
\caption{Selected crossed-evaluation alignment contrasts for 17 and 18 models. Primary matched contrasts use 27 blocks and 2,159 items with at least three options; the supplementary full-set contrast uses 30 blocks and 2,399 items. Identifier minus text denotes label scoring minus text scoring with labels (L0--LT0); printed minus unprinted denotes text scoring with labels minus text scoring without labels (LT0--U0). Intervals are conditional 95\% paired block-bootstrap intervals from 10,000 draws, with fitted maps held fixed.}
\label{tab:s-crossed-alignment}
\begin{tabular}{@{}lllrr@{}}
\toprule
Models & Blocks & Contrast & Estimate & 95\% interval \\
\midrule
17 & 27 & Identifier minus text & 0.0630 & [0.0365, 0.0945] \\
17 & 27 & Printed minus unprinted & -0.0192 & [-0.0419, 0.0011] \\
17 & 30 & Printed minus unprinted & -0.0288 & [-0.0521, -0.0073] \\
18 & 27 & Identifier minus text & 0.0908 & [0.0534, 0.1338] \\
18 & 27 & Printed minus unprinted & -0.0227 & [-0.0446, -0.0036] \\
18 & 30 & Printed minus unprinted & -0.0273 & [-0.0472, -0.0087] \\
\bottomrule
\end{tabular}
\end{table}

\begin{table}[htbp]
\centering\small
\caption{Between-fit agreement of label-scoring ranks before and after fitting identifier effects (L0.$s$) with text-scoring ranks without labels (U0.$s$). Ranks are computed over 2,399 items. Shared-17 ranks from the 18-model fits are recomputed among the common models. Mean distance is the mean absolute difference in rank; U0 is an alternative evaluation reference.}
\label{tab:s-crossed-ranks}
\begin{tabular}{@{}lrrrrrr@{}}
\toprule
Fit / ranked set & $\rho$ before & $\rho$ after & Distance before & Distance after & Closer & Farther \\
\midrule
17 / 17 & 0.9804 & 0.8824 & 0.588 & 1.765 & 1 & 9 \\
18 / 17 & 0.9804 & 0.8824 & 0.588 & 1.765 & 1 & 9 \\
18 / 18 & 0.9835 & 0.9009 & 0.556 & 1.667 & 1 & 9 \\
\bottomrule
\end{tabular}
\end{table}

\begin{table}[htbp]
\centering\small
\caption{Numerical acceptance of retained fits. The acceptance threshold is a maximum absolute gradient of $10^{-4}$, alongside finite parameters, independently recomputed objectives and gradients, and input identity checks. The BBH rows are the three selected display fits; crossed counts include both SVD and reciprocal starts. Acceptance establishes first-order stationarity under these checks.}
\label{tab:s-numerical-acceptance}
\begin{tabular}{@{}lrrr@{}}
\toprule
Analysis & Fits & Accepted & Largest absolute gradient \\
\midrule
BBH M0 & 1 & 1 & 6.83e-06 \\
BBH M1 & 1 & 1 & 9.83e-06 \\
BBH M2 & 1 & 1 & 9.19e-06 \\
Held-out BBH & 20 & 20 & 9.88e-06 \\
Crossed evaluation & 16 & 16 & 9.93e-06 \\
\bottomrule
\end{tabular}
\end{table}

\begin{table}[htbp]
\centering\small
\caption{Starting-solution sensitivity within each anchor-plus-four-perturbation set. Each specification has five numerically accepted candidates. Minimum rank correlation and maximum movement cover both $s$ and $q$ across all pairs; alignment covers all five answer families. These descriptive ranges are not confidence intervals. Component values can vary even when ranks and alignment remain close.}
\label{tab:s-start-stability}
\begin{tabular}{@{}lrrrr@{}}
\toprule
Specification & Candidates & Minimum rank $\rho$ & Maximum move & Maximum family change \\
\midrule
M1 & 5 & 0.9999755 & 13 & 0.009063 \\
M2 & 5 & 0.9999997 & 4 & 0.000998 \\
\bottomrule
\end{tabular}
\end{table}

\begin{table}[htbp]
\centering\small
\caption{Component and rank ranges for the three ranking examples across each specification's five saved local candidates. Mean interaction, mean offset and $s$ use logit units. Ranges include the preceding anchor and four perturbations; they describe starting-solution sensitivity, not sampling uncertainty.}
\label{tab:s-example-start-ranges}
\begin{tabular}{@{}llrrrr@{}}
\toprule
Specification & Model & $\bar I$ & $\bar\delta$ & $s$ & Rank $s$ \\
\midrule
M1 & Phi & [-2.708, -2.693] & [0.836, 0.838] & [-3.323, -3.312] & [814, 816] \\
M1 & DeepSeek & [-4.806, -4.795] & [-1.302, -1.286] & [-0.926, -0.920] & [97, 97] \\
M1 & QwQ & [-2.748, -2.735] & [-0.076, -0.074] & [-2.486, -2.478] & [441, 444] \\
M2 & Phi & [-2.474, -2.473] & [0.176, 0.176] & [-2.219, -2.218] & [615, 615] \\
M2 & DeepSeek & [-2.682, -2.682] & [-0.710, -0.710] & [-1.213, -1.213] & [229, 229] \\
M2 & QwQ & [-2.548, -2.547] & [-0.178, -0.177] & [-1.870, -1.869] & [434, 434] \\
\bottomrule
\end{tabular}
\end{table}

\clearpage\small\setlength{\LTcapwidth}{\linewidth}
\begin{longtable}{@{}p{.35\linewidth}rrrrrr@{}}
\caption{All 24 archived BBH task profiles for Phi (P) and DeepSeek (D) in the selected M1 fit. Accuracy is observed correctness. Offsets ($\delta$) and relative interactions ($R$) are task means in logit units; $R$ subtracts the model-specific mean interaction over all items. Task-relative interaction therefore describes the remaining fitted profile after the explicit answer component is represented. Display names Navigation and Adjective order correspond to the archived identifiers navigate and hyperbaton.}\label{tab:s-task-profiles}\\
\toprule
Task & Acc. P & Acc. D & $\delta_P$ & $\delta_D$ & $R_P$ & $R_D$ \\
\midrule\endfirsthead
\toprule
Task & Acc. P & Acc. D & $\delta_P$ & $\delta_D$ & $R_P$ & $R_D$ \\
\midrule\endhead
boolean expressions & 0.804 & 0.868 & -0.019 & -1.519 & 0.208 & 0.294 \\
causal judgement & 0.615 & 0.519 & -0.115 & -1.933 & 0.484 & 0.094 \\
date understanding & 0.560 & 0.656 & 1.219 & -1.341 & 0.174 & 0.238 \\
disambiguation qa & 0.716 & 0.396 & 1.186 & -1.429 & 0.207 & -0.177 \\
formal fallacies & 0.624 & 0.652 & 0.468 & -1.360 & 0.197 & -0.121 \\
geometric shapes & 0.216 & 0.556 & 0.240 & -0.803 & 0.542 & 0.306 \\
Adjective order (Hyperbaton) & 0.920 & 0.560 & 1.537 & -2.127 & -1.931 & -2.288 \\
Logical deduction (5) & 0.552 & 0.504 & 1.254 & -1.202 & 0.116 & 0.056 \\
Logical deduction (7) & 0.512 & 0.496 & 1.174 & -0.951 & 0.323 & 0.318 \\
Logical deduction (3) & 0.828 & 0.548 & 1.139 & -1.306 & -0.279 & 0.486 \\
movie recommendation & 0.624 & 0.728 & 1.168 & -0.947 & 0.617 & 0.483 \\
Navigation & 0.636 & 0.420 & 0.007 & -2.531 & 0.785 & 0.219 \\
object counting & 0.392 & 0.612 & 0.741 & 1.389 & 0.787 & 0.151 \\
penguins in a table & 0.582 & 0.596 & 1.238 & -1.116 & -0.093 & 0.211 \\
reasoning about colored objects & 0.488 & 0.700 & 0.383 & -1.138 & 0.087 & 1.169 \\
ruin names & 0.652 & 0.864 & 1.246 & -1.039 & -0.082 & 0.936 \\
salient translation error detection & 0.568 & 0.616 & 1.214 & -1.325 & -0.029 & 0.507 \\
snarks & 0.787 & 0.657 & 1.541 & -2.159 & -0.850 & -0.798 \\
sports understanding & 0.592 & 0.484 & -0.042 & -2.289 & 0.921 & 0.294 \\
temporal sequences & 0.680 & 0.888 & 1.190 & -0.940 & -1.851 & 0.144 \\
Tracking objects (5) & 0.252 & 0.172 & 1.234 & -1.196 & -0.181 & -0.754 \\
Tracking objects (7) & 0.228 & 0.148 & 1.166 & -0.879 & -0.043 & -0.392 \\
Tracking objects (3) & 0.320 & 0.332 & 1.145 & -1.324 & -0.340 & -1.118 \\
web of lies & 0.488 & 0.552 & -0.077 & -2.119 & 0.070 & -0.378 \\
\bottomrule\end{longtable}\normalsize

\clearpage
\begin{figure}[p]
\centering
\includegraphics[width=\linewidth,height=.81\textheight,keepaspectratio] {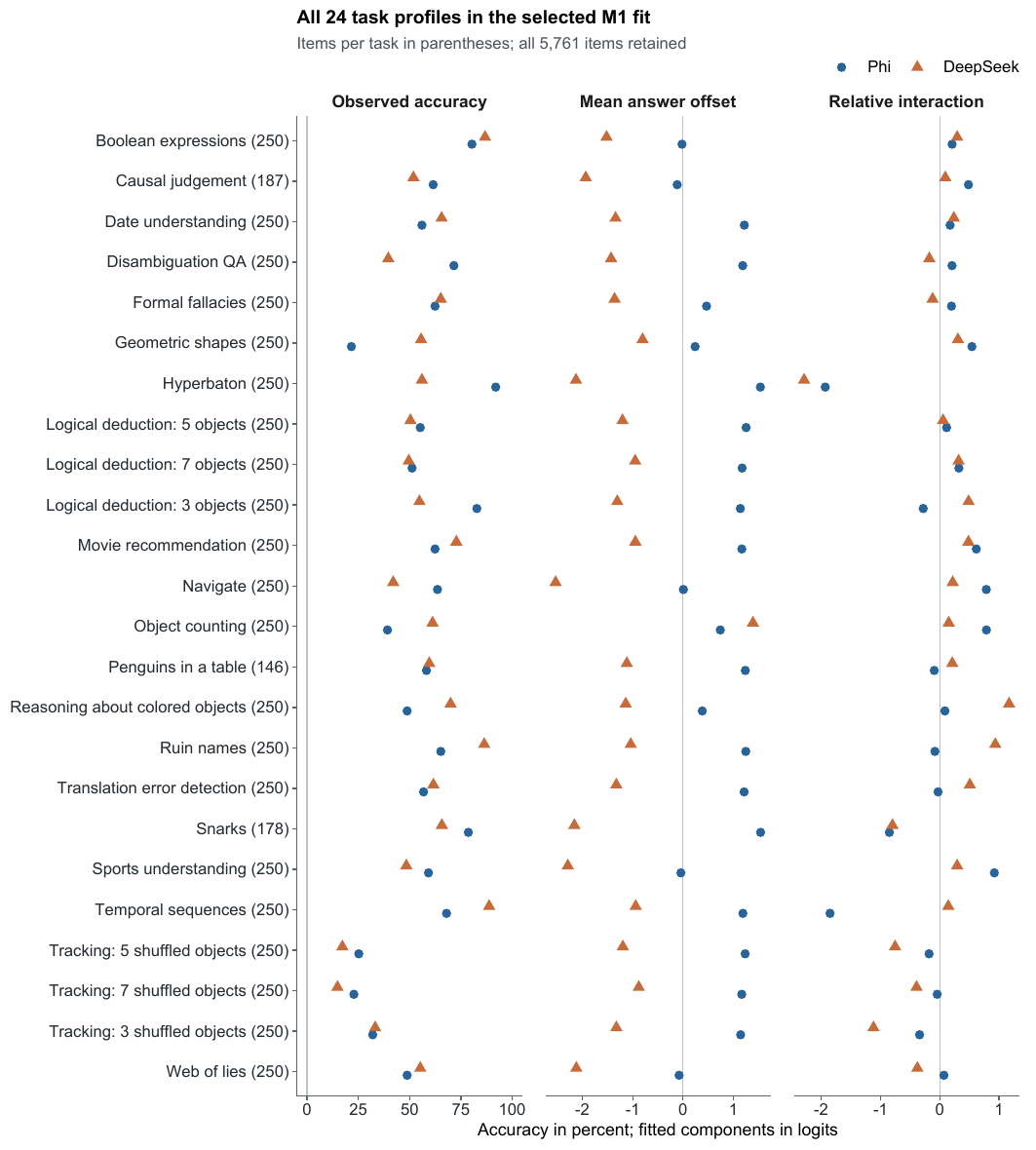}
\caption[Complete task profiles]{\input{notes/figureS1_caption}}
\label{fig:task-profiles}
\end{figure}

\end{document}